\documentclass[11pt,a4paper]{article}
\usepackage[hyperref]{acl2021}
\usepackage{times}
\usepackage{latexsym}

\usepackage{microtype}

\usepackage{graphicx}
\usepackage{multirow}
\usepackage{amsmath}
\usepackage{amsfonts}
\usepackage{bm}
\usepackage{booktabs}
\usepackage{subfigure}
\usepackage{verbatim}
\usepackage[ruled,vlined,linesnumbered]{algorithm2e} 
\usepackage{amssymb}

\usepackage{stfloats}
\aclfinalcopy 

\title{Multi-Label Few-Shot Learning for Aspect Category Detection}

\author{Mengting Hu\textsuperscript{1} \quad Shiwan Zhao\textsuperscript{2}\thanks{\; Shiwan Zhao is the corresponding author.} \quad Honglei Guo\textsuperscript{2} \quad Chao Xue\textsuperscript{3}\thanks{\; The work was (partially) done in IBM.} \\ {\bf Hang Gao\textsuperscript{1} \quad Tiegang Gao\textsuperscript{1} \quad Renhong Cheng\textsuperscript{1} \quad Zhong Su\textsuperscript{2}} \\
\textsuperscript{1} Nankai University \quad \textsuperscript{2} IBM Research - China \quad \textsuperscript{3} JD Explore Academy \\
\{mthu, knimet\}@mail.nankai.edu.cn, \{zhaosw, guohl, suzhong\}@cn.ibm.com \\
xuechao19@jd.com, \{gaotiegang, chengrh\}@nankai.edu.cn
}

\date{}

\begin{document}
\maketitle
\begin{abstract}
Aspect category detection (ACD) in sentiment analysis aims to identify the aspect categories mentioned in a sentence. In this paper, we formulate ACD in the few-shot learning scenario. However, existing few-shot learning approaches mainly focus on single-label predictions. These methods can not work well for the ACD task since a sentence may contain multiple aspect categories. Therefore, we propose a multi-label few-shot learning method based on the prototypical network. To alleviate the noise, we design two effective attention mechanisms. The support-set attention aims to extract better prototypes by removing irrelevant aspects. The query-set attention computes multiple prototype-specific representations for each query instance, which are then used to compute accurate distances with the corresponding prototypes. To achieve multi-label inference, we further learn a dynamic threshold per instance by a policy network. Extensive experimental results on three datasets demonstrate that the proposed method significantly outperforms strong baselines.
\end{abstract}

\section{Introduction}
Aspect category detection (ACD) \cite{pontiki-etal-2014-semeval,pontiki-etal-2015-semeval} is an important task in sentiment analysis. It aims to identify the aspect categories mentioned in a given sentence from a predefined set of aspect categories. For example, in the sentence \emph{``the cheesecake is tasty and the staffs are friendly''}, two aspect categories, i.e. \emph{food} and \emph{service}, are mentioned. The performance of existing approaches for the ACD task \cite{Zhou2015Representation,Schouten2018Supervised,hu-etal-2019-constrained} relies heavily on the scale of the labeled dataset. They usually suffer from limited data and fail to generalize well to novel aspect categories with only a few labeled instances. On the one hand, it is time-consuming and labor-intensive to annotate large-scale datasets. On the other hand, given a large dataset, many long-tail aspects still suffer from data sparsity. 


\begin{figure}[t]
\centering
\includegraphics[width=0.480\textwidth]{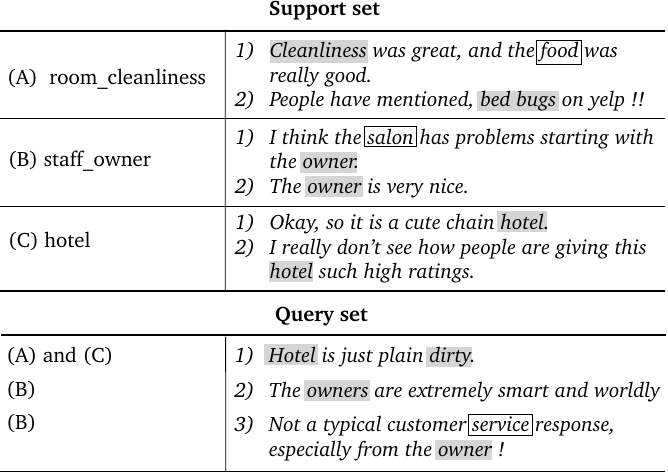}
\caption{Example meta-task in a 3-way 2-shot scenario. The words in gray background describe the target aspects of interest, while the words marked by the rectangle are irrelevant aspects, which tend to be noise for this meta-task.}
\label{figure_example}
\end{figure}

Few-shot learning (FSL) provides a solution to address the above challenges. FSL learns like a human, identifying novel classes with limited supervised information by exploiting prior knowledge. Many efforts have been devoted to FSL \cite{ravi2017iclr,finn2017model,snell2017prototypical,wang2018low,gao2019hybrid}. Among these methods, the prototypical network \cite{snell2017prototypical} is a promising approach, which is simple but effective. It follows the meta-learning paradigm by building a collection of $N$-way $K$-shot meta-tasks. A meta-task aims to infer a query set with the help of a small labeled support set. It first learns a prototype for each class in the support set. Then the query instance is predicted by measuring the distance with $N$ prototypes in the embedding space. 


In this paper, we formulate ACD in the FSL scenario, which aims to detect aspect categories accurately with limited training instances. However, ACD is a multi-label classification problem since a sentence may contain multiple aspect categories. Most FSL works learn a single-label classifier and can not work well to address the ACD task. The reasons are two-fold. Firstly, the sentences of each class (i.e., aspect category) in the support set are diverse and contain noise from irrelevant aspects. As displayed in Figure \ref{figure_example}, there are three classes in the support set, and each class has two instances. The aspect categories \emph{food} and \emph{salon} tend to be noise for this meta-task, making it hard to learn a good prototype for each class in the support set. Secondly, the query set is also noisy. Figure \ref{figure_example} demonstrates three different cases. The first sentence mentions two aspects \emph{hotel} and \emph{room\_cleanliness} out of the support set. We need to detect both aspects accurately as multi-label classification. When detecting each of them, the other aspect acts as noise and makes the task hard. The second sentence is an easy case with a single aspect \emph{staff\_owner}. The third sentence mentions the aspect \emph{staff\_owner} out of the support set, while the aspect \emph{service} is noise for this meta-task. In summary, the noise from both the support set and query set makes the few-shot ACD a challenging task. 

To this end, we propose a multi-label FSL method based on the prototypical network \cite{snell2017prototypical}. We alleviate the noise in the support set and query set by two effective attention mechanisms. Concretely, the support-set attention tries to extract the common aspect of each class. By removing the noise (i.e., irrelevant aspects), the support-set attention can yield better prototypes. Then for a query instance, the query-set attention utilizes the prototypes to compute multiple prototype-specific query representations, in which the irrelevant aspects are removed. Given the better prototypes and the corresponding prototype-specific query representations, we can compute accurate distances between the query instance and the prototypes in the embedding space. We detect the aspect categories in the query instance by ranking the distances. To select the positive aspects from the ranking, we design a policy network \cite{williams1992simple} to learn a dynamic threshold for each instance. The threshold is modeled as the action of the policy network with continuous action space.

The main contributions of our work are as follows: 
\begin{itemize}
    \item We formulate ACD as a multi-label FSL problem and design a multi-label FSL method based on the prototypical network to solve the problem. To the best of our knowledge, we are the first to address ACD in the few-shot scenario. 
    \item To alleviate the noise from the support set and query set, we design two effective attention mechanisms, i.e., support-set attention and query-set attention.
    \item Experimental results on the three datasets demonstrate that our method outperforms strong baselines significantly.
\end{itemize}


\begin{figure*}[t]
\centering
\includegraphics[width=0.990\textwidth]{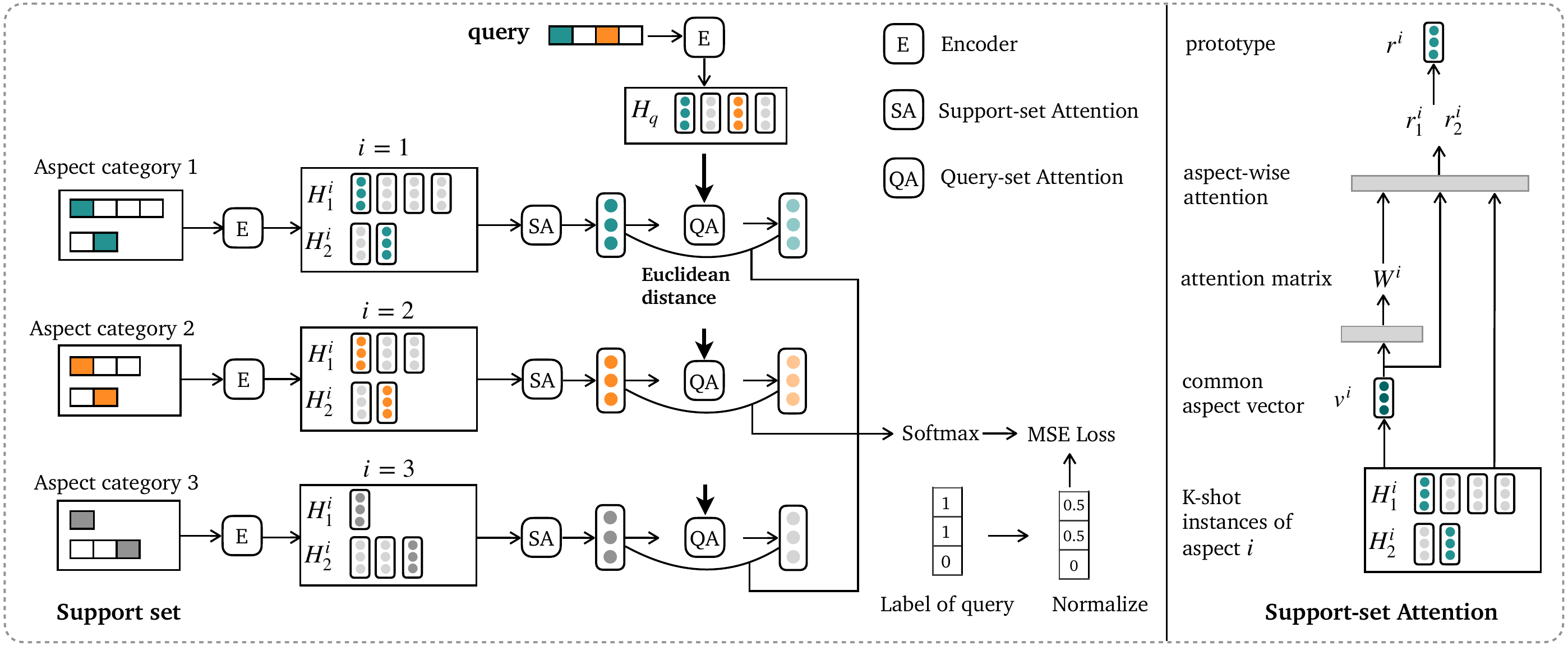}
\caption{The left part depicts the main network for an example $N$-way $K$-shot meta-task with a query instance  ($N=3,K=2$). Each small cube of the instance symbolizes an aspect category. The colored cubes indicate the target aspects of interest while the white cubes indicate the noisy aspects. The right part shows the details of the support-set attention.}
\label{network}
\end{figure*}

\section{Related Work}
\textbf{Aspect Category Detection} \; Previous works for ACD can mainly be divided into two types: unsupervised and supervised methods. Unsupervised approaches extract aspects by mining semantic association \cite{su2006using} or co-occurrence frequency \cite{hai2011implicit,Schouten2018Supervised}. These methods require a large corpus to mine aspect knowledge and have limited performance. Supervised methods address this task via hand-crafted features \cite{kiritchenko2014nrc}, automatically learning useful representations \cite{Zhou2015Representation}, multi-task learning \cite{xue-etal-2017-mtna,hu-etal-2019-constrained}, or topic-attention model \cite{movahedi2019aspect}. The above methods detect aspect categories out of a pre-defined set, which cannot handle the unseen classes. These challenges motivate us to investigate this task in the few-shot scenario.


\noindent
\textbf{Few-Shot Learning} \; Few-shot learning (FSL) \cite{fe2003bayesian,fei2006one} is close to real artificial intelligence, which borrows the learning process from the human. By incorporating the prior knowledge, it obtains new knowledge fast with limited supervised information. Many works have been proposed for FSL, which can be mainly divided into four research directions. 

One promising direction is distance-based methods. These methods measure the distance between instances in the feature embedding space. The siamese network \cite{koch2015siamese} infers the similarity score between an instance pair. Others compare the cosine similarity \cite{vinyals2016matching} or Euclidean distance \cite{snell2017prototypical}. The relation network \cite{sung2018learning} exploits a neural network to learn the distance metric. Afterward, \newcite{garcia2017few} utilize graph convolution network to extract the structural information of classes. The second direction focuses on the optimization of networks. Model-agnostic meta-learning (MAML) algorithm \cite{finn2017model} learns a good initialization of the model and updates the model by a few labeled examples. Meta networks \cite{munkhdalai2017meta} achieve rapid generalization via fast parameterization. The third type is based on hallucination \cite{wang2018low,li2020adversarial}. This research line directly deals with data deficiency by ``learning to augment'', which designs a generator on the base classes and then hallucinates novel class data to augment few-shot samples. The last direction introduces a weight generator to predict classification weight given a few novel class samples, either based on attention mechanism \cite{gidaris2018dynamic} or Gaussian distribution \cite{guo2020attentive}.


A recent work Proto-HATT \cite{gao2019hybrid} is similar to ours. Proto-HATT is based on the prototypical network \cite{snell2017prototypical}, which deals with the text noise in the relation classification task by employing hybrid attention at both the instance-level and the feature-level. This method is designed for single-label FSL. Compared with it, our method designs two attention mechanisms to alleviate the noise on the support set and query set, respectively. The collaboration of two attentions helps compute accurate distances between the query instance and prototypes, and then improves multi-label FSL. 


\noindent
\textbf{Multi-Label Few-Shot Learning} \; Compared with single-label FSL, the multi-label FSL has been underexplored. Previous works focus on image synthesis \cite{alfassy2019laso} and signal processing \cite{cheng2019multi}. \newcite{rios-kavuluru-2018-shot} develop few-shot and zero-shot methods for multi-label text classification when there is a known structure over the label space. Their approach relies on label descriptors and the hierarchical structure of the label spaces, which limits its application in practice. \newcite{hou2020few} propose to address the multi-label intent detection task in the FSL scenario. It calibrates the threshold by kernel regression. Different from this work, we learn a dynamic threshold per instance in a reinforced manner.

\section{Methodology}

In the few-shot ACD scenario, each meta-task contains a support set $S$ and a query set $Q$. The meta-task is to assign the query instance to the class(es) of the support set. An instance may be a multi-aspect sentence. Thus a query sentence may describe more than one class out of the support set\footnote{We found that the probability of a query instance belonging to more than one class is around 4.5\% in the ACD dataset, i.e. FewAsp, by randomly sampling 10,000 5-way 5-shot meta-tasks with 5 query sentences for each class.}. Therefore, we define the few-shot ACD as a multi-label few-shot classification problem. 

\subsection{Overview}
Suppose in an $N$-way $K$-shot meta-task, the support set is $S=\{(x_1^i,...x_K^i), y^i\}_{i=1}^N$, where each $x^i$ is a sentence and $(x_1^i,...,x_K^i)$ all contain the aspect category $y^i$. A query instance is $(x_q,\bm{y_q})$, where $\bm{y_q}$ is a binary label vector indicating the aspects in $x_q$ out of $N$ classes. 

Figure \ref{network} presents the main network by an example 3-way 2-shot meta-task. It is composed of three modules, i.e., encoder, support-set attention (SA) and query-set attention (QA). Each class in the support set contains $K$ instances, which are fed into the encoder to obtain $K$ encoded sequences. Next, SA module extracts a prototype for this class from the encoded sequences. After obtaining $N$ prototypes, we feed a query instance into the QA module to compute multiple prototype-specific query representations, which are then used to compute the Euclidean distances with the corresponding prototypes. Finally, we normalize the negative distances to obtain the ranking of prototypes and then select the positive predictions (i.e., aspect categories) by a dynamic threshold. Next, we will introduce the modules of our method in detail.

\subsection{Encoder}
Given an input sentence $x=\{w_1,w_2,...,w_n\}$, we first map it into an embedding sequence $\{\bm{e_1},\bm{e_2},...,\bm{e_n}\}$ by looking up the pre-trained GloVe embeddings \cite{pennington2014glove}. Then we encode the embedding sequence by a convolutional neural network (CNN) \cite{zeng-etal-2014-relation,gao2019hybrid}. The convolution kernel slides with the window size $m$ over the embedding sequence. We gain the contextual sequence $H=\{\bm{h_1},\bm{h_2},...,\bm{h_n}\}$,  $H\in{\mathbb{R}^{n\times{d}}}$:
\begin{equation}
    \bm{h_i} = \mathrm{CNN}(\bm{e}_{i-\frac{m-1}{2}},...,\bm{e}_{i+\frac{m-1}{2}})
\end{equation}
where CNN($\cdot$) is a convolution operation. The advantages of CNN are two-fold: first, the convolution kernel can extract n-gram features on the receptive field. For example, the bi-gram feature of \emph{hot dog} could help detect the aspect category \emph{food}; second, CNN enables parallel computing over inputs, which is more efficient \cite{xue-li-2018-aspect}.

\subsection{Support-set Attention (SA)}
In each class of the support set, the $K$-shot instances describe a common aspect, i.e., the target aspect of interest\footnote{In almost all cases, there is only one common aspect in the $K$ instances. We randomly sample 10,000 5-way 5-shot meta-tasks, and found that the probability of containing more than one common aspect in each class is less than 0.086\%. The probability will be much lower in the 10-way scenario.}. As shown in Figure \ref{figure_example}, two sentences, \emph{``Cleanliness was great, and the food was really good''} and \emph{``People have mentioned, bed bugs on yelp!!''}, share the common aspect \emph{room\_cleanliness}. The former contains two aspect categories \emph{room\_cleanliness} and \emph{food}. In this example meta-task, it is an instance of the class \emph{room\_cleanliness}. However, when sampling other meta-tasks, the instance may be used to represent the class \emph{food}. This leads to confusion and makes learning a good prototype difficult. To deal with the issue brought by multi-aspect sentences, we first need to identify the common aspect. As depicted in the right part of Figure \ref{network}, we compute the \emph{common aspect vector} by the combination of the $K$-shot instances. We then regard the vector as a condition and inject it into the attention mechanism to make our attention mechanism aspect-wise. 

\noindent
\textbf{Common Aspect Vector} \; The encoded $K$-shot instances of a class contain one common aspect and some irrelevant aspects. Among these aspects, the common aspect is the majority. Thus, we simply conduct a word-level average to extract the common aspect vector $\bm{v^i}\in{\mathbb{R}^{d}}$.
\begin{equation}
    \bm{v^i} = \mathrm{avg}(H_1^i,H_2^i,...,H_K^i)
\end{equation}

The average operation highlights the common aspect, but cannot completely eliminate noisy aspects. 
To further reduce the noise of irrelevant aspects in each instance, we use the common aspect as the condition in the attention mechanism. 

\noindent
\textbf{Aspect-Wise Attention} \; To make the attention mechanism adapt to the condition, we have two designs. First, we directly use the common aspect vector to compute the attention with each instance (see Eq. \ref{eq:support_att}), which filters out the irrelevant aspects of each instance to some extent. Second, we exploit the idea of dynamic conditional network, which has been demonstrated effective in FSL \cite{zhao2018dynamic}. By predicting a dynamic attention matrix with the common aspect vector, our attention mechanism can further adapt to the condition, i.e., the common aspect vector of the class. Specifically, we learn different perspectives of the condition by simply repeating the common aspect vector \cite{vaswani2017attention}. Then it is fed into a linear layer to obtain the attention matrix $W^i$ for class $i$. 
\begin{equation}
    W^i = W(\bm{v^i}\otimes{e_M}) + \bm{b}
\label{equation:repeat_vi}
\end{equation}
where $(\bm{v^i}\otimes{e_M})\in\mathbb{R}^{e_M\times{d}}$ is the operation repeatedly concatenating $\bm{v^i}$ for $e_M$ times. The linear layer has parameter matrix $W\in\mathbb{R}^{d\times{e_M}}$ and bias $\bm{b}\in\mathbb{R}^{d}$. This layer is shared in the classes of all meta-tasks, which is learned to be class-agnostic. Thus in the testing phase, it can generate aspect-wise attention for a novel class. 

Then in class $i$ of the support set, we exploit the common aspect vector and attention matrix to calculate a denoised representation for every instance. The denoised representation $\bm{r^i_j}$ for the $j$-th instance is computed as below.
\begin{equation}
\begin{split}
    \bm{\beta} &= \mathrm{softmax}(\bm{v^i}\mathrm{tanh}(H^i_jW^i)) \\
    \bm{r^i_j} &= \bm{\beta}H^i_j
\end{split}
\label{eq:support_att}
\end{equation}

In this way, the support-set attention is adapted to the condition and is also class-specific. Thus it tends to focus on the correct aspect even for a multi-aspect sentence representing different classes. 

Finally, the average of denoised representations for $K$-shot instances is the prototype of this class.
\begin{equation}
    \bm{r^i} = \mathrm{avg}(\bm{r^i_1}, \bm{r^i_2}, ..., \bm{r^i_K})
\end{equation}

After processing all classes in the support set, we obtain $N$ prototypes $\{\bm{r^1}, \bm{r^2}, ..., \bm{r^N}\}$.

\subsection{Query-set Attention (QA)}
A query instance may also contain multiple aspects, making the sentence noisy. To deal with the noise in a query instance, we select the relevant aspects from the query instance by the QA module. Specifically, we first process the query instance by the encoder and obtain the encoded instance $H_q$. Then we feed $H_q$ into the QA module to obtain multiple prototype-specific query representations $\bm{r^i_q}$ by the $N$ prototypes.
\begin{equation}
\begin{split}
    \bm{\rho^i} &= \mathrm{softmax}(\bm{r^i}\mathrm{tanh}(H_q)) \\
    \bm{r^i_q} &= \bm{\rho^i}H_q
\end{split}
\label{eq:query_attention}
\end{equation}

The QA module tries to focus on the aspect category which is similar to the prototype. In Eq. \ref{eq:query_attention}, the attention is non-parametric. It can reduce the dependence on parameters and can accelerate the adaptation to unseen classes. 

\subsection{Training Objective}
For a query instance, we compute the Euclidean distance (ED) between each prototype and its prototype-specific query representation, and we obtain $N$ distances. Next, we normalize the negative distances as the final prediction, which is a ranking of the prototypes.
\begin{equation}
\begin{aligned}
    \bm{\hat{y}} = \mathrm{softmax}(-\mathrm{ED}(\bm{r^i},\bm{r^i_q})) & , & i\in{[1,N]}
\end{aligned}
\label{eq:ranking}
\end{equation}

The training objective is the mean square error (MSE) loss:
\begin{equation}
    L = \sum{(\bm{\hat{y}} - \bm{y_q})}^2
\label{eq:loss_mse}
\end{equation}
where $\bm{y_q}$ is the ground-truth. We also normalize $\bm{y_q}$ to ensure the consistency between the prediction and the ground-truth. 

\noindent
{\bf Learning Dynamic Threshold (DT)} \; To select the positive aspects from the ranking (see Eq. \ref{eq:ranking}) for a query instance, we further learn a dynamic threshold. The threshold is modeled by a policy network \cite{williams1992simple}, which has a continuous action space following Beta distribution \cite{chou2017improving}. Given a query instance, we define the \emph{state} as $[(\bm{r^1}-\bm{r^1_q})^2;...;(\bm{r^N}-\bm{r^N_q})^2;\hat{\bm{y}}]$. We feed the state into the policy network and obtain the parameters $a$ and $b$ of a Beta distribution. Then we sample a threshold $\tau$ from $Beta(\tau|a,b)$. The \emph{reward} $score$ is the F1 score for this instance based on $\tau$. We also introduce a reference $score^*$, which is the F1 score based on a baseline action, i.e., the mode of $Beta(\tau|a,b)$: $\frac{a-1}{a+b-2}$. The training objective is defined as below to minimize the negative expected reward.
\begin{equation}
    L_t=-(score-score^*)\mathrm{log}P(\tau)
\label{eq:loss_policy}
\end{equation}
where $P(\tau)$ is the probability of $\tau$ in the Beta distribution. During inference, we select the positive aspects in $\bm{\hat{y}}$ with the baseline action.

\section{Experiments}

\begin{table}[t!]
\small
\begin{center}
\begin{tabular} {|l|ccc|}
\hline
     {\bf Dataset} & {\bf \#cls.} & {\bf \#inst./cls.} & {\bf \#inst.}  \\
     \hline
     FewAsp(single) & 100 & 200 & 20000 \\
     FewAsp(multi)  & 100 & 400 & 40000 \\
     FewAsp   & 100 & 630 & 63000 \\
\hline
\end{tabular}
\end{center}
\caption{\label{dataset} Statistics of three datasets. \#cls. denotes the number of classes. \#inst./cls. denotes the number of instances per class. \#inst. denotes the total number of instances.}
\end{table}

\begin{table*}[t!]
\small
\begin{center}
\begin{tabular} {|l|cc|cc|cc|cc|}
\hline
    \multirow{2}{*}{Models} &
    \multicolumn{2}{c|}{5-way 5-shot} & \multicolumn{2}{c|}{ 5-way 10-shot} & \multicolumn{2}{c|}{10-way 5-shot} & \multicolumn{2}{c|}{10-way 10-shot} \\
    & AUC & F1 & AUC & F1 & AUC & F1 & AUC & F1  \\
    \hline
    Relation Network     & 0.9331 & 75.79 & 0.9086 & 72.02 & 0.9181 & 63.78 & 0.9054 & 61.15 \\
    Matching Network     & 0.9705 & 81.89 & 0.9749 & 84.62 & 0.9630 & 70.95 & 0.9672 & 73.28 \\
    Graph Network        & 0.9654 & 81.45 & 0.9746 & 85.04 & 0.9545 & 70.75 & 0.9697 & 77.84 \\
    Prototypical Network & 0.9649 & 83.30 & 0.9753 & 86.29 & 0.9597 & 74.23 & 0.9671 & 76.83 \\
    IMP & 0.9665 & 83.69 & 0.9747 & 86.14 & 0.9600 & 73.80 & 0.9691 & 77.09 \\
    Proto-HATT           & 0.9645 & 83.33 & 0.9762 & 86.71 & 0.9571 & 73.42 & 0.9700 & 77.65 \\
    \hline
    Proto-AWATT (ours) & {\bf 0.9756}$^{\dagger\ddagger}$ & {\bf 86.71}$^{\dagger\ddagger}$ & {\bf 0.9796} & {\bf 88.54}$^{\dagger\ddagger}$ & {\bf 0.9701}$^{\dagger\ddagger}$ & {\bf 80.28}$^{\dagger\ddagger}$ & {\bf 0.9755}$^{\dagger\ddagger}$ & {\bf 82.97}$^{\dagger\ddagger}$ \\
\hline
\end{tabular}
\end{center}
\caption{\label{table-result-single} Evaluation results in terms of AUC and macro-f1 (\%) on FewAsp(single). All results are the average of 5 runs. The marker $^\dagger$ refers to $p$-value$<$0.05 of the T-test when comparing with Prototypical Network. The marker $^\ddagger$ refers to $p$-value$<$0.05 when comparing with Proto-HATT.}
\end{table*}

\begin{table*}[t!]
\small
\begin{center}
\begin{tabular} {|l|cc|cc|cc|cc|}
\hline
    \multirow{2}{*}{Models} &
    \multicolumn{2}{c|}{5-way 5-shot} & \multicolumn{2}{c|}{ 5-way 10-shot} & \multicolumn{2}{c|}{10-way 5-shot} & \multicolumn{2}{c|}{10-way 10-shot} \\
    & AUC & F1 & AUC & F1 & AUC & F1 & AUC & F1  \\
    \hline
    Relation Network     & 0.8491 & 58.38 & 0.8621 & 61.37 & 0.8422 & 43.71 & 0.8472 & 44.85 \\
    Matching Network     & 0.8954 & 65.70 & 0.9138 & 69.02 & 0.8828 & 50.86 & 0.8994 & 54.42 \\
    Graph Network        & 0.8797 & 59.25 & 0.9045 & 64.63 & 0.8605 & 45.42 & 0.8844 & 48.49 \\
    Prototypical Network & 0.8967 & 67.88 & 0.9160 & 72.32 & 0.8801 & 52.72 & 0.9068 & 58.92 \\
    IMP & 0.9012 & 68.86 & 0.9229 & 73.51 & 0.8871 & 53.96 & 0.9110 & 59.86 \\
    Proto-HATT           & 0.9110 & 69.15 & 0.9303 & 73.91 & {\bf 0.9044} & 55.34 & {\bf 0.9238} & 60.21 \\
    \hline
    Proto-AWATT (ours) & {\bf 0.9145}$^\dagger$ & {\bf 71.72}$^\dagger$ & {\bf 0.9389}$^\dagger$ & {\bf 77.19}$^{\dagger\ddagger}$ &  {0.8980}$^\dagger$ & {\bf 58.89}$^{\dagger\ddagger}$ & {0.9234}$^\dagger$ & {\bf 66.76}$^{\dagger\ddagger}$ \\
\hline
\end{tabular}
\end{center}
\caption{\label{table-result-multi} Evaluation results in terms of AUC and macro-f1 (\%) on FewAsp(multi).}
\end{table*}


\subsection{Datasets}
We construct three few-shot ACD datasets from Yelp\_aspect \cite{10.1145/3097983.3098170}, which is a large-scale multi-domain dataset for aspect recommendation. We group all instances by aspects and choose 100 aspect categories. Following \newcite{han-etal-2018-fewrel}, we split the 100 aspects without intersection into 64 aspects for training, 16 aspects for validation, and 20 aspects for testing.

According to the sentence type, i.e., single-aspect or multi-aspect\footnote{A sentence contains a single aspect or multiple aspects.}, we sample different types of sentences from each group and construct three datasets: FewAsp(single), FewAsp(multi), and FewAsp, which are composed of single-aspect, multi-aspect, and both types of sentences, respectively. Note that FewAsp is randomly sampled from the original set of each class, which can better reflect the data distribution in real applications. The statistics of the three datasets are shown in Table \ref{dataset}.

\subsection{Experimental Settings}
{\bf Evaluation Metrics} \; Previous single-label FSL \cite{snell2017prototypical} usually evaluates performance by accuracy. In the multi-label setting, we choose AUC (Area Under Curve) and macro-f1 as the evaluation metrics. AUC is utilized for model selection and macro-f1 is computed with a threshold. In our experiments, we found that for all methods in three datasets, the overall best thresholds are 0.3 in the 5-way setting and 0.2 in the 10-way setting. Thus we choose them for evaluating the baselines. 

\noindent
{\bf Training Details} \; We first train the main network with MSE loss $L$ (Eq. \ref{eq:loss_mse}). Then we initialize the main network with the learned parameters and jointly train the policy network with $L_t$ (Eq. \ref{eq:loss_policy}). The implementation details are described in the appendix.

\begin{table*}[t!]
\small
\begin{center}
\begin{tabular} {|l|cc|cc|cc|cc|}
\hline
    \multirow{2}{*}{Models} &
    \multicolumn{2}{c|}{5-way 5-shot} & \multicolumn{2}{c|}{ 5-way 10-shot} & \multicolumn{2}{c|}{10-way 5-shot} & \multicolumn{2}{c|}{10-way 10-shot} \\
    & AUC & F1 & AUC & F1 & AUC & F1 & AUC & F1  \\
    \hline
    Relation Network     & 0.8556 & 59.52 & 0.8698 & 62.78 & 0.8494 & 45.62 & 0.8377 & 44.70 \\
    Matching Network     & 0.9076 & 67.14 & 0.9239 & 70.09 & 0.8844 & 51.27 & 0.8990 & 54.61 \\
    Graph Network        & 0.8948 & 61.49 & 0.9235 & 69.89 & 0.8735 & 47.91 & 0.9019 & 56.06 \\
    Prototypical Network & 0.8888 & 66.96 & 0.9177 & 73.27 & 0.8735 & 52.06 & 0.9013 & 59.03 \\
    IMP & 0.8995 & 68.96 & 0.9230 & 74.13 & 0.8850 & 54.14 & 0.9081 & 59.84 \\
    Proto-HATT           & 0.9154 & 70.26 & 0.9343 & 75.24 & 0.9063 & 57.26 & 0.9286 & 61.51 \\
    \hline
    Proto-AWATT (ours) & {\bf 0.9335}$^{\dagger\ddagger}$ & {\bf 75.37}$^{\dagger\ddagger}$ & {\bf 0.9528}$^{\dagger\ddagger}$ & {\bf 80.16}$^{\dagger\ddagger}$ & {\bf 0.9206}$^\dagger$ & {\bf 65.65}$^{\dagger\ddagger}$ & {\bf 0.9342}$^\dagger$ & {\bf 69.70}$^{\dagger\ddagger}$ \\
\hline
\end{tabular}
\end{center}
\caption{\label{table-result-full} Evaluation results in terms of AUC and macro-f1 (\%) on FewAsp.}
\end{table*}

\begin{table}[t!]
\small
\begin{center}
\begin{tabular} {|l|cc|}
\hline
    \multirow{2}{*}{Models} & \multicolumn{2}{c|}{FewAsp} \\
    & AUC & F1 \\
    \hline
    Proto-AWATT (ours) & {\bf 0.9206} & {\bf 65.65} \\
    \hline
    w/o SA      & 0.8890 & 54.34 \\
    w/o attention matrix $W^i$  & 0.9128 & 61.68 \\
    w/o QA  & 0.8886 & 51.19 \\
    \hline
    w/o DT & 0.9161 & 64.48 \\
    w/o DT w/ KR & 0.9159 & 64.06 \\
    w/o DT w/ MS & 0.9163 & 64.00 \\
\hline
\end{tabular}
\end{center}
\caption{\label{table-result-ablation} Ablation study of the 10-way 5-shot scenario on FewAsp.}
\end{table}


\subsection{Compared Methods}
Our approach is named as {\bf Proto-AWATT} (aspect-wise attention). We validate the effectiveness of the proposed method by comparing with the following popular approaches. 
\begin{itemize}
    \vspace{3pt}
    \item {\bf Matching Network} \cite{vinyals2016matching}: It is a metric-based attention method, where distance is measured by cosine similarity.
    
    \vspace{3pt}
    \item {\bf Prototypical Network} \cite{snell2017prototypical}: It computes the average of embedded support examples for each class as the prototype, and then measures the distance between the embedded query instance and each prototype.
    
    \vspace{3pt}
    \item {\bf Relation Network} \cite{sung2018learning}: It utilizes a neural network to learn the relation metric. 
    
    \vspace{3pt}
    \item {\bf Graph Network} \cite{garcia2017few}: It casts FSL as a supervised message passing task by graph neural network.
    
    \vspace{3pt}
    \item {\bf IMP} \cite{pmlr-v97-allen19b}: It proposes infinite mixture prototypes to represent each class by a set of clusters, with the number of clusters determined directly from the data.
    
    \vspace{3pt}
    \item {\bf Proto-HATT} \cite{gao2019hybrid}: It is based on the prototypical network, which deals with the noise with hybrid instance-level and feature-level attention mechanisms. 
    
\end{itemize}

\subsection{Experimental Analysis}
We report the experimental results of various methods in Table \ref{table-result-single}, Table \ref{table-result-multi}, Table \ref{table-result-full} and Table \ref{table-result-ablation}. The best scores on each metric are marked in bold. The experimental results demonstrate the effectiveness of our method.

\noindent
{\bf Overall Performance} \; AUC and macro-f1 scores of all the methods are shown in Table \ref{table-result-single}, Table \ref{table-result-multi} and Table \ref{table-result-full}. Firstly, we observe that our method Proto-AWATT achieves the best results on almost all evaluation metrics of the three datasets. This reveals the effectiveness of the proposed method. 
Secondly, compared to Proto-HATT, Proto-AWATT achieves significant improvement. It is worth noting that the average improvement of macro-f1 on three datasets is 4.99\%. This exhibits that the SA and QA modules successfully reduce noise for few-shot ACD. Meanwhile, accurate distance measurement between prototypes and the prototype-specific query representations can facilitate the detection of multiple aspects in the query instance.

Then we found that all methods on FewAsp(multi) perform consistently worse than the counterparts on FewAsp(single) and FewAsp. This is because more aspects increase the complexity of the dataset. On FewAsp(multi), Proto-AWATT still outperforms other methods in most settings, which demonstrates the robustness of our model on various data distributions.

In general, the 10-way scenario contains much more noise than the 5-way. We observe that compared to Proto-HATT, Proto-AWATT achieves more significant improvements in the 10-way scenario than the 5-way. The results further indicate that Proto-AWATT can really alleviate the noise. 


\begin{table}[t!]
\small
\begin{center}
\begin{tabular} {|l|cc|cc|}
\hline
    \multirow{2}{*}{Models} & \multicolumn{2}{c|}{Proto-HATT} & \multicolumn{2}{c|}{Proto-AWATT} \\
    & AUC & F1 & AUC & F1 \\
    \hline
    GloVe + CNN & 0.9063 & 57.26 & 0.9206 & 65.65 \\
    GloVe + LSTM & 0.9137 & 59.46 & 0.9357 & 66.86 \\
    BERT & 0.8971 & 57.33 & 0.9459 & 70.09 \\
    DistilBERT & 0.9067 & 59.57 & 0.9451 & 70.23 \\
\hline
\end{tabular}
\end{center}
\caption{\label{table-result-ablation-encoder} Ablation study of using different encoders in the 10-way 5-shot scenario on FewAsp.}
\end{table}

\noindent
{\bf Ablation Study} \; Table \ref{table-result-ablation} depicts the results of ablation study. Firstly, without the SA module, the performances of Proto-AWATT drop a lot. In particular, AUC drops by 3.43\%, and macro-f1 drops by 17.23\% relatively. This verifies that the SA module helps reduce noise and extract better prototypes. We can also see that without attention matrix $W^i$ in SA causes consistent decreases on all metrics. This suggests that predicting dynamic attention matrix for each class is effective, which makes the SA module extract better prototypes. Then we found that without the QA module, Proto-AWATT significantly performs worse. This validates that for a query instance, computing multiple prototype-specific query representations helps obtain accurate distances for ranking, which facilitates the multi-label predictions. 

Finally, when removing DT and using a static threshold ($\tau=0.2$ in the 10-way setting), it causes a slight decrease. This shows that learning dynamic threshold is effective. We further compare DT with two alternative dynamic threshold methods: (1) MS (mean $\pm$ standard deviation of the threshold by cross-validation); (2) a kernel regression (KR) approach which is proposed by \newcite{hou2020few} to calibrate the threshold. Comparing with MS and KR, our method slightly outperforms them. This is because DT benefits from reinforcement learning and directly optimizes the evaluation metrics.


\noindent
{\bf Different Encoders} \; We also compare the performances of our method with a strong baseline Proto-HATT when using different encoders to obtain the contextual sequence $H$. The results are reported in Table \ref{table-result-ablation-encoder}. The output of pre-trained encoders, i.e., BERT \cite{devlin2018bert} or DistilBERT \cite{sanh2019distilbert}, are directly used as the contextual sequence. We observe that Proto-AWATT significantly outperforms the strong baseline Proto-HATT on all encoders. 

\begin{figure}[t]
\centering
\includegraphics[width=0.44\textwidth]{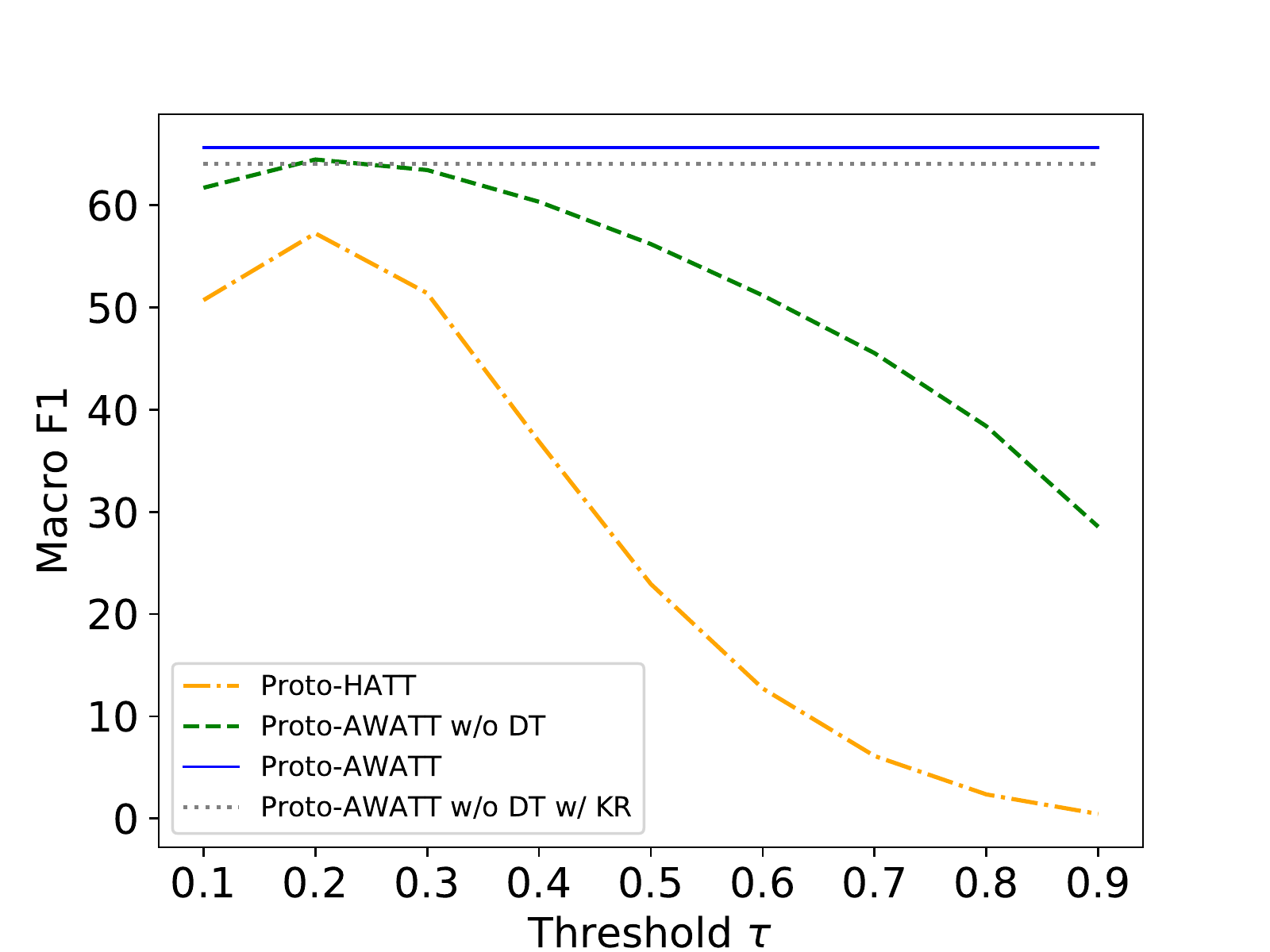}
\caption{Macro-f1 scores for different thresholds on 10-way 5-shot setting of FewAsp.}
\label{figure-threshold_10}
\end{figure}

\noindent
{\bf Effects of Thresholds} \; As depicted in Figure \ref{figure-threshold_10}, we analyze the impact of different thresholds on the macro-f1 score during inference. 
We can see that Proto-AWATT without DT consistently outperforms Proto-HATT in various thresholds. Macro-f1 scores of the two methods are getting worse as $\tau$ grows. However, the declines in Proto-HATT are more significant. At $\tau=0.9$, the macro-f1 of Proto-HATT drops nearly to 0. Proto-AWATT without DT still achieves much higher macro-f1. This indicates that the proposed two attention mechanisms help extract an accurate ranking of prototypes. The ranking is less sensitive to the threshold, which makes our method robust and stable. We also found that learning threshold by DT benefits from a reinforced way, which slightly outperforms KR and the best static threshold.


\begin{figure}[t]
\centering
\subfigure[Proto-HATT]{
\includegraphics[width=0.220\textwidth]{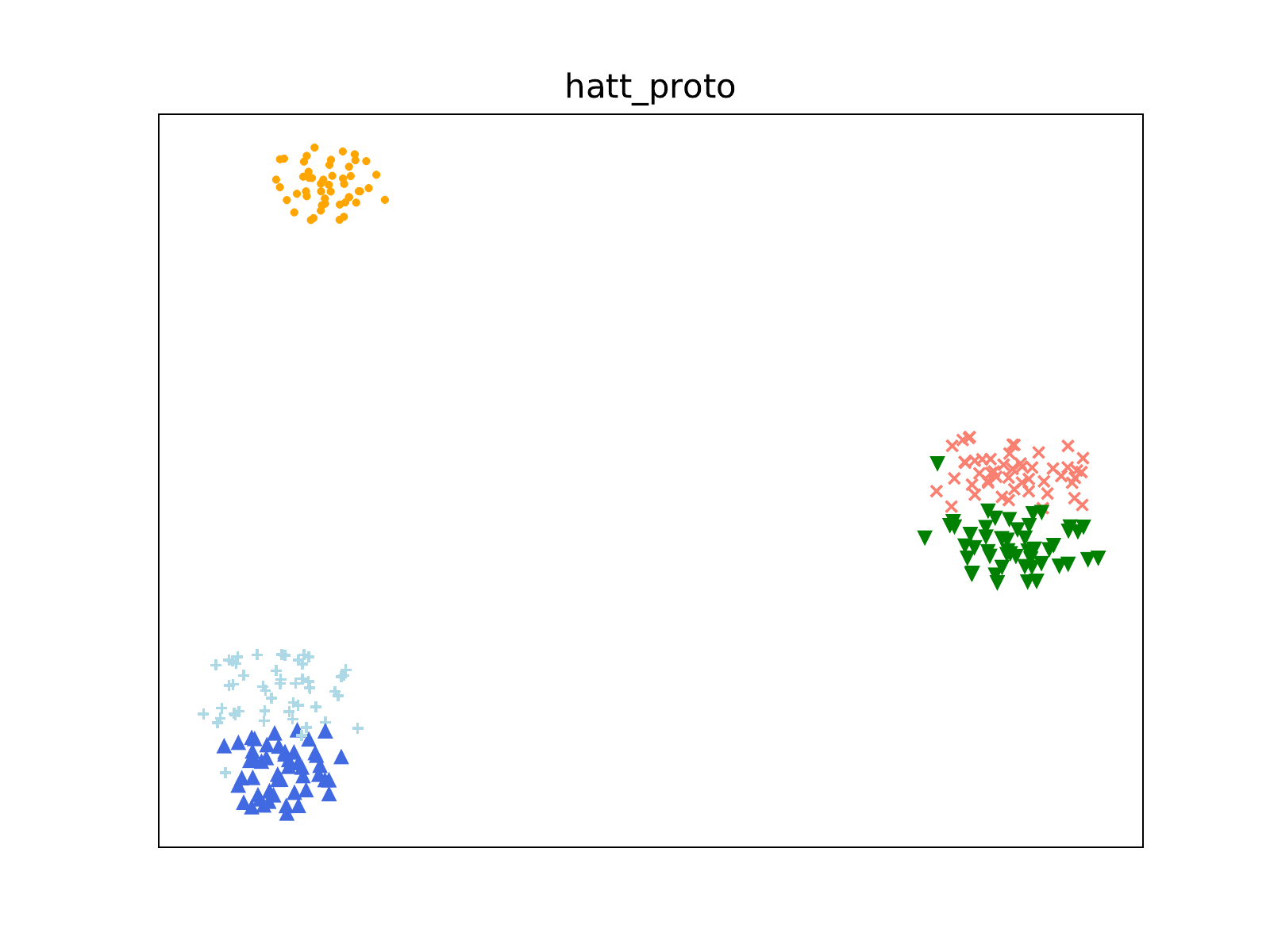}}
\subfigure[Proto-AWATT]{
\includegraphics[width=0.220\textwidth]{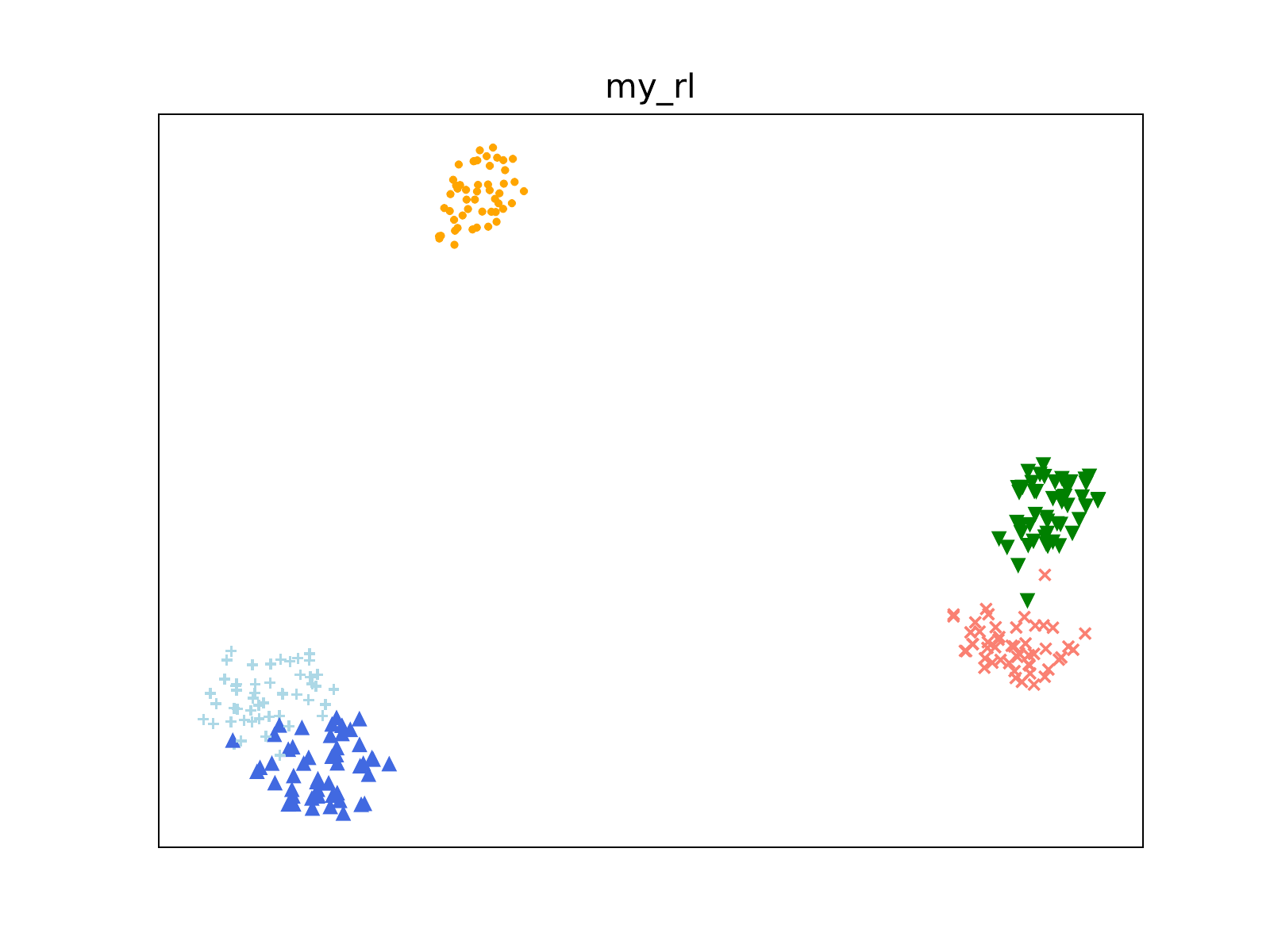}}
\caption{Visualization of extracted prototypes for the support set.}
\label{figure-tsne-support}
\end{figure}

\begin{figure}[t]
\centering
\subfigure[Proto-HATT]{
\includegraphics[width=0.220\textwidth]{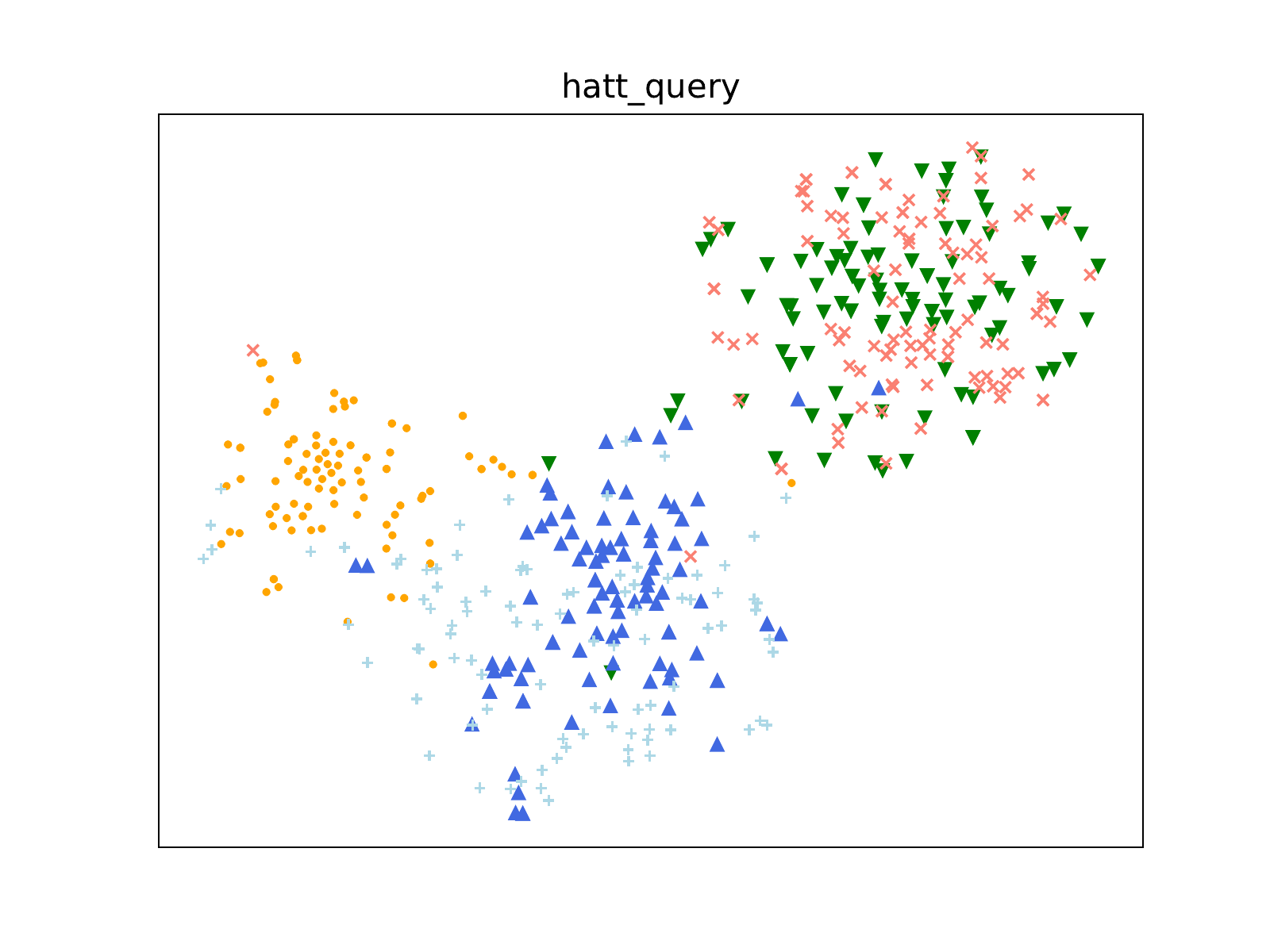}}
\subfigure[Proto-AWATT]{
\includegraphics[width=0.220\textwidth]{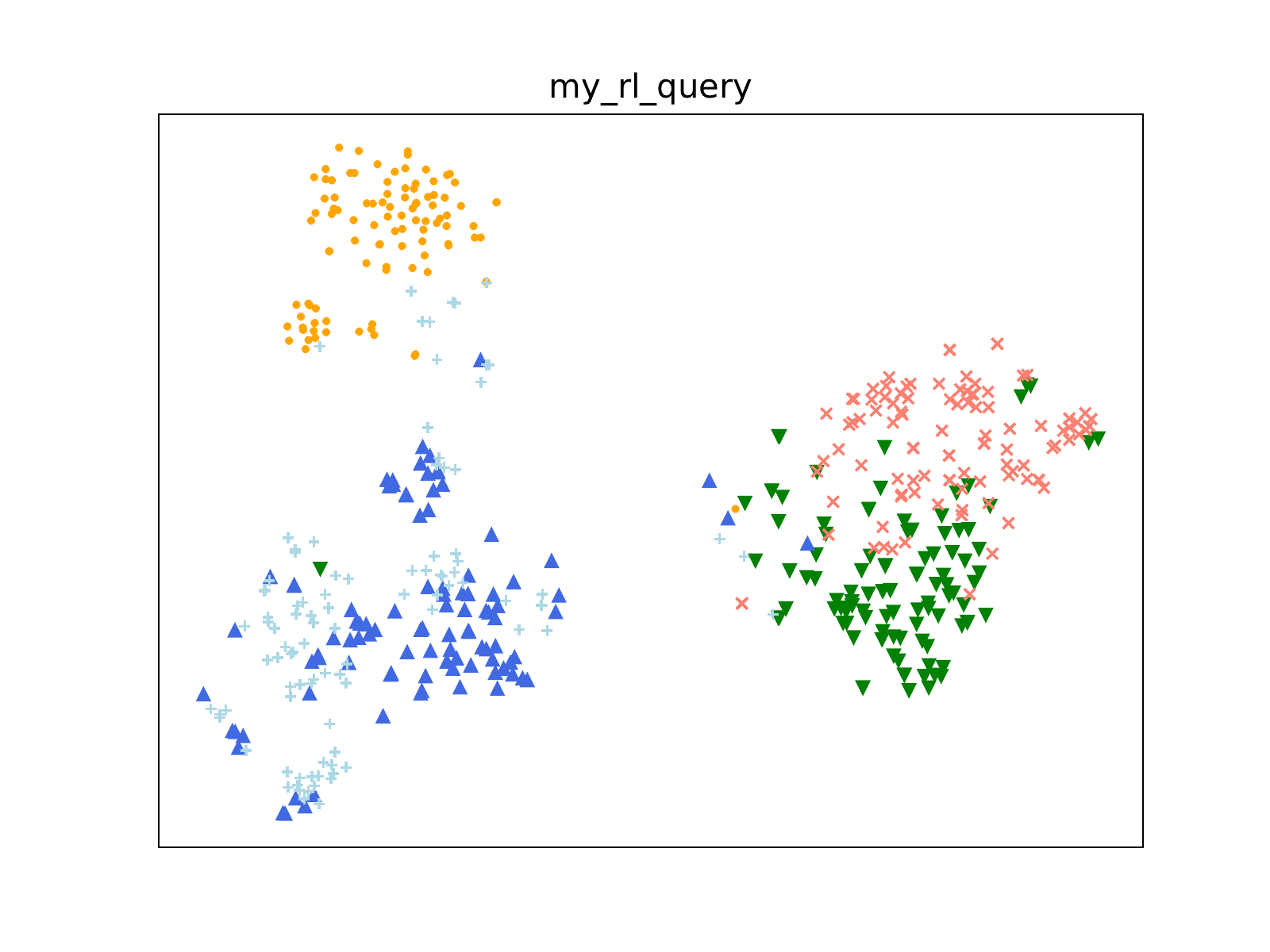}}
\caption{Visualization of extracted representations for the query set.}
\label{figure-tsne-query}
\end{figure}

\subsection{Visualizations}
We further analyze Proto-AWATT by visualizing the extracted representations from the support set and query set, respectively. The representations are visualized by t-SNE \cite{maaten2008visualizing}. To observe the performance in a challenging situation, we choose the testing set from {\bf FewAsp(multi)} as an example.

\noindent
{\bf Support Set} \; Figure \ref{figure-tsne-support} presents the visualization of extracted prototypes from two methods. We randomly sample 5 classes and then sample 50 times of 5-way 5-shot meta-tasks for the five classes. Then for each class, we have 50 prototype vectors. We observe that prototype vectors from our approach are more separable than those from Proto-HATT. This further indicates that the SA module can alleviate noise and thus yield better prototypes.

\noindent
{\bf Query Set} \; We randomly sample 5 classes and then sample 20 times of 5-way 5-shot meta-tasks for these classes. Each meta-task has 5 query instances per class. Thus we have $25\times20=500$ query instances. It is worth noting that our model learns $N$ prototype-specific query representations for each query instance. We choose the representations according to the ground-truth label. However, Proto-HATT only outputs a single representation for a query instance. As depicted in Figure \ref{figure-tsne-query}, we can see that the representations learned by our method are obviously more separable than those by Proto-HATT. This further reveals that Proto-AWATT can obtain accurate prototype-specific query representations, which contributes to computing accurate distances.

\section{Conclusion}
In this paper, we formulate the aspect category detection (ACD) task in the few-shot learning (FSL) scenario. Existing FSL methods mainly focus on single-label predictions. They can not work well for the ACD task since a sentence may contain multiple aspect categories. Therefore, we propose a multi-label FSL method based on the prototypical network. Specifically, we design two effective attention mechanisms for the support set and query set to alleviate the noise from both sets. To achieve multi-label inference, we further learn a dynamic threshold per instance by a policy network with continuous action space. Extensive experimental results in three datasets demonstrate that our method outperforms strong baselines significantly.

\section*{Acknowledgements}
We sincerely thank all the anonymous reviewers for providing valuable feedback. This work is supported by the National Science and Technology Major Project, China (Grant No. 2018YFB0204304).

\bibliography{acl2021}
\bibliographystyle{acl_natbib}

\clearpage
\begin{appendix}
\begin{table*}[t!]
\small
\begin{center}
\setlength{\tabcolsep}{3.0mm}{
\begin{tabular} {|l|cc|cc|cc|}
\hline
    \multirow{2}{*}{Models} &
    \multicolumn{2}{c|}{FewAsp(single)} & \multicolumn{2}{c|}{FewAsp(multi)} & \multicolumn{2}{c|}{FewAsp} \\
    & AUC & F1 & AUC & F1 & AUC & F1  \\
    \hline
    Proto-AWATT (ours) & 0.9701 & {\bf 80.28} & 0.8980 & 58.89 & {\bf 0.9206} & {\bf 65.65} \\
    \hline
    w/o SA     & 0.9304 & 63.63 & 0.8854 & 53.02 & 0.8890 & 54.34 \\
    w/o attention matrix $W^i$  & {\bf 0.9703} & 78.08 & 0.8959 & 57.42 & 0.9128 & 61.68 \\
    w/o QA     & 0.9541 & 69.61 & 0.8920 & 51.51 & 0.8886 & 51.19 \\
\hline
    w/o DT & 0.9689 & 79.46 & 0.8970 & {\bf 59.40} & 0.9161 & 64.48 \\
    w/o DT w/ KR & 0.9695 & 79.41 & {\bf 0.9006} & 59.13 & 0.9159 & 64.06 \\
    w/o DT w/ MS & 0.9681 & 78.73 & 0.8976 & 59.01 & 0.9163 & 64.00 \\
\hline
\end{tabular}}
\end{center}
\caption{\label{table-result-ablation-appendix} Ablation study of the 10-way 5-shot scenario on three datasets.}
\end{table*}

\section{Implementation Details} 
\textbf{Hyperparameters} \; All baselines and our model are implemented by Pytorch. We initialize word embeddings with 50-dimension GloVe vectors and fine-tune them during the training. All other parameters are initialized by sampling from a normal distribution $\mathcal{N}(0,0.1)$. The dimension of the hidden state $d$ is 50. The convolutional window size $m$ is set as 3. The optimizer is Adam with a learning rate $10^{-3}$. When jointly training the policy network, the learning rate is set to $10^{-4}$. In each dataset, we construct four FSL tasks, where $N=5, 10$ and $K=5, 10$. And the number of query instances per class is 5. For example, in a 5-way 10-shot meta-task, there are $5\times10=50$ instances in the support set and $5\times5=25$ instances in the query set.

\noindent
\textbf{Dynamic Threshold (DT)} \; In this module, we first map the state into a vector representation through linear layers. Then the vector is mapped into two separate linear layers with softplus as the activation function. We obtain the parameters of Beta distribution, i.e. $a$ and $b$, respectively. When training the policy network, a reward is computed based on the $\mathrm{softmax}$ output (i.e. ranking of prototypes). However, the $\mathrm{softmax}$ output is narrow and highly confident, resulting in sparse rewards. Therefore, we exploit a temperature $T=2$ to make the $\mathrm{softmax}$ output more smooth. In addition, two-stage training is also designed to deal with the sparse rewards. We first train the main network to obtain accurate rankings. Then when learning the policy network, we can gain more meaningful rewards.

\noindent
\textbf{Training Details} \; In every epoch, we randomly sample 800 meta-tasks for training. The number of meta-tasks during validation and testing are both set as 600. The average score of meta-tasks are used for evaluation. We employ an early stop strategy if the AUC score of the validation set is not improved in 3 epochs, and the best model is chosen for testing. For all baselines and our model, we report the average testing results from 5 runs, where the seeds are set to [5, 10, 15, 20, 25]. All models are trained on one Tesla P100 GPU with 16GB of RAM.


\section{Experimental Results}
\textbf{Ablation Study} \; We display the results of ablation study on three datasets in Table \ref{table-result-ablation-appendix}. 

\begin{figure}[t]
\centering
\includegraphics[width=0.45\textwidth]{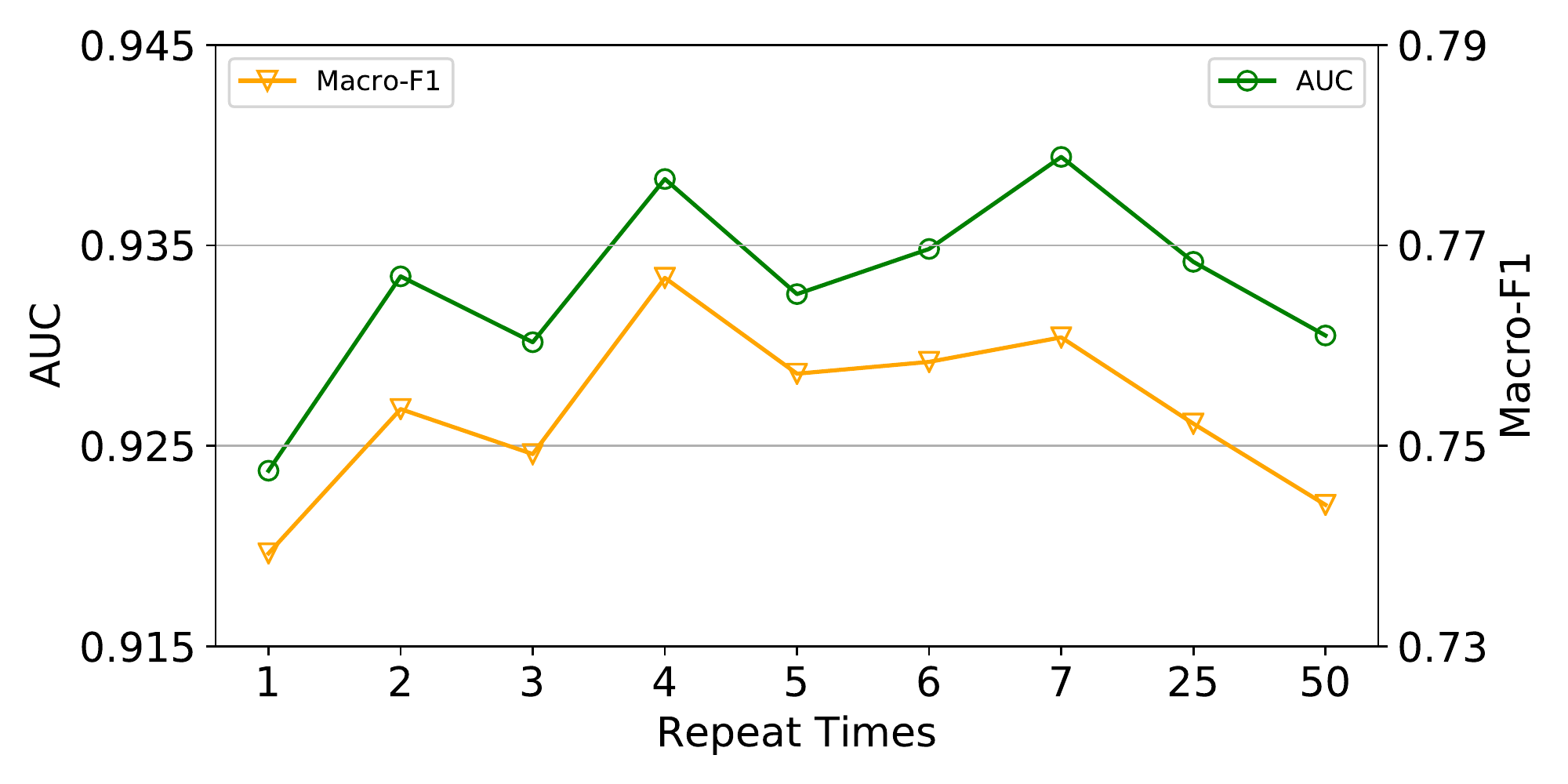}
\caption{Effects of attention matrix on 5-way 5-shot setting of FewAsp.}
\label{fig:repeat}
\end{figure}

\noindent
\textbf{Effects of Attention Matrix} \; To explore the effects of the condition on the attention matrix, we compare the performances of Proto-AWATT by setting different repeat times $e_M$ in Eq. \ref{equation:repeat_vi}. The results are displayed in Figure \ref{fig:repeat}. We can see that by repeating more times of the common aspect vector, the AUC and macro-f1 score both outperform the results of setting $e_M=1$. As $e_M$ grows, the performances are improved. However, when setting $e_M$ as 25 or even 50, the performances decline. A possible reason is that the model tends to overfit the training classes.

\end{appendix}

\end{document}